\documentclass{article} % ENSURE 11pt is here

\usepackage{makecell}
\usepackage{amsfonts}
\usepackage{amssymb}
\usepackage{enumitem} % For itemized lists

\usepackage{spconf,amsmath,graphicx}

\usepackage{enumitem}
\setlist{nosep, leftmargin=14pt}

\usepackage{mwe} % to get dummy images
\title{MSDM: Generating Task-Specific Pathology Images with a Multimodal Conditioned Diffusion Model for Cell and Nuclei Segmentation}
\name{Dominik Winter, Mai Bui, Monica Azqueta Gavaldon, Nicolas Triltsch, Marco Rosati, Nicolas Brieu}
\address{\small AstraZeneca Computational Pathology GmbH, Landsberger Str. 300, 80687 Munich, Germany}
\begin{document}
\ninept
\maketitle

\begin{abstract}
  Scarcity of annotated data, particularly for rare or atypical morphologies, present significant challenges for cell and nuclei segmentation in computational pathology. While manual annotation is labor-intensive and costly, synthetic data offers a cost-effective alternative. 
  We introduce a Multimodal Semantic Diffusion Model (MSDM) for generating realistic pixel-precise image-mask pairs for cell and nuclei segmentation.
  By conditioning the generative process with cellular/nuclear morphologies (using horizontal and vertical maps), RGB color characteristics, and BERT-encoded assay/indication metadata, MSDM generates datasests with desired morphological properties. 
  These heterogeneous modalities are integrated via multi-head cross-attention, enabling fine-grained control over the generated images.
  Quantitative analysis demonstrates that synthetic images closely match real data, with low Wasserstein distances between embeddings of generated and real images under matching biological conditions. The incorporation of these synthetic samples, exemplified by columnar cells, significantly improves segmentation model accuracy on columnar cells.
  This strategy systematically enriches data sets, directly targeting model deficiencies. We highlight the effectiveness of multimodal diffusion-based augmentation for advancing the robustness and generalizability of cell and nuclei segmentation models. Thereby, we pave the way for broader application of generative models in computational pathology.
\end{abstract}

\section{Introduction}
\label{sec:intro}

\begin{figure}[htb]
  \begin{minipage}[b]{1.0\linewidth}
    \centering
    \centerline{\includegraphics[width=8.5cm]{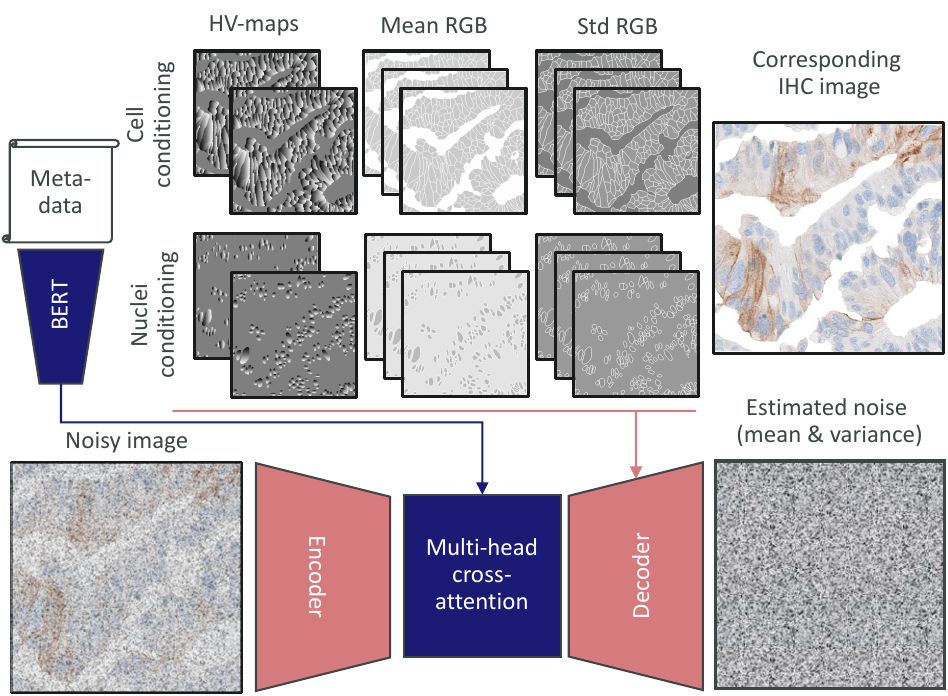}}
    \vspace*{-2mm}
    \caption{
      MSDM overview: We condition the model on cell and nuclei morphologies using horizontal and vertical (HV) maps, and on RGB color characteristics (distinguishing foreground/background pixels). Additionally, BERT-encoded textual information conditions the diffusion model via a multi-head cross-attention mechanism to generate synthetic images that match the morphology of existing segmentation masks, creating a new image-mask pair for model training.
    }
    \vspace*{-2mm}
  \label{figure1}
\end{minipage}
\end{figure}

High-quality datasets, traditionally requiring labor-intensive and costly expert annotations \cite{van2021deep}, are crucial for deep learning in computational pathology, particularly for fundamental tasks like cell and nuclei segmentation upon which diagnosis, prognosis, and biomarker discovery are often based.
While specialized methods such as Cellpose \cite{stringer2021cellpose} generally deliver convincing results, their performance on specific morphologies, particularly rare ones, often suffers from a lack of high-quality training data. 
Synthetic data offers an efficient, cost-effective alternative to create realistic high-fidelity datasets \cite{kazerouni2023diffusion}.
In this study, we introduce MSDM, a Multimodal Semantic Diffusion Model, for generating synthetic immunohistochemistry (IHC) images. MSDM's strong conditioning capabilities enable the generation of data with desired morphological properties, directly addressing deficiencies in existing segmentation models for challenging morphologies or morphologies with low prevalence. 
Our goal is to generate realistic, pixel-precise image-mask pairs by re-using existing cell and nuclei annotations, ensuring the realism of the generated images. 
Unlike natural images with distinct objects, pathological images contain numerous, small, often touching cell instances requiring fine-grained spatial conditioning, aspects not fully addressed by previous methods for this domain \cite{wang2022semantic,xu2025topocellgen}. 
We showcase our approach's capabilities to improve segmentation models' performance on low prevalence morphologies, exemplified on columnar cells. Columnar cells, characterized by large eccentricity (as defined in section \ref{sec:enhancement}), are underrepresented in our initial training dataset, leading to suboptimal performance of the segmentation model in our test set. 
Adding images of columnar cells generated with MSDM boosts the performance of segmentation models.
\begin{figure}[htb]
  \begin{minipage}[b]{1.0\linewidth}
    \centering
    \centerline{\includegraphics[width=8.5cm]{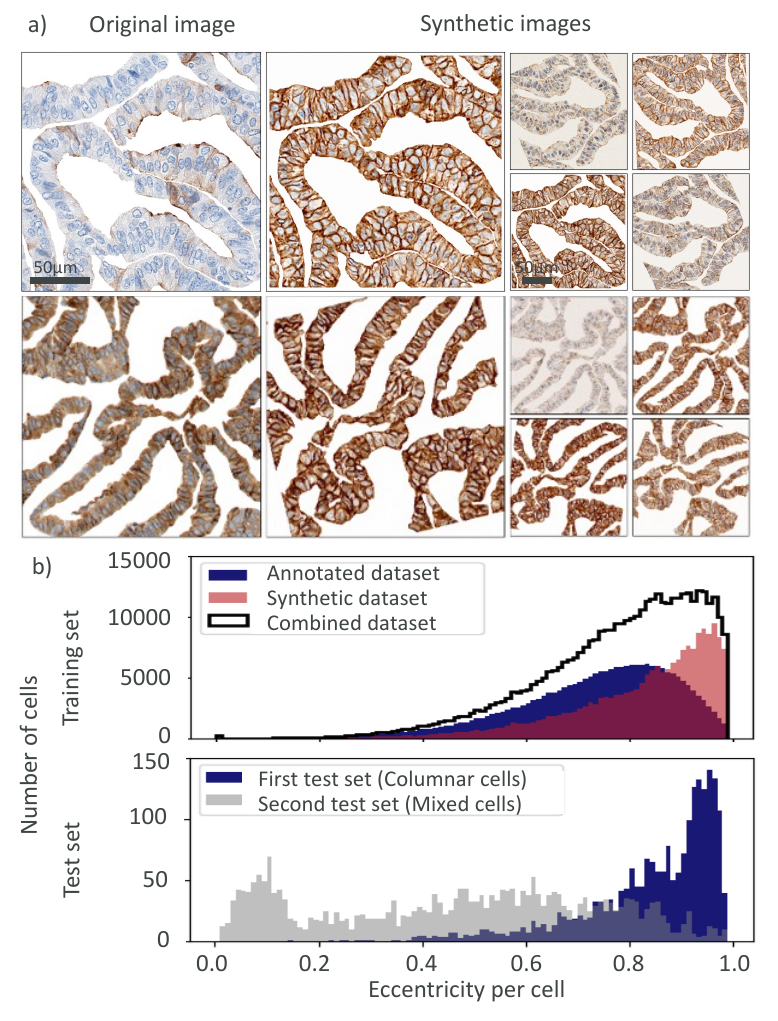}}
    \vspace*{-2mm}
    \caption{
      \textbf{a} Original and corresponding synthetic images, for visualization images with the same (rotated) target mask are shown \textbf{b} Distribution of cell eccentricities in the training-set before and after data augmentation with MSDM and in the two test-sets.
    }
    \vspace*{-2mm}
  \label{figure2}
\end{minipage}
\end{figure}
\section{Background}
\label{sec:background}
Synthetic data offers a cost-effective alternative to labor-intensive expert annotations. Early image-to-image translation (I2I) models, such as Pix2Pix, relied on CNNs and adversarial training to transfer images between domains while preserving content. These approaches have been successfully adapted in computational pathology for domain translation \cite{brieu2024auxiliary} and training data generation \cite{winter2024restaingan}, enabling the reuse of existing annotations.
Denoising diffusion probabilistic models (DDPMs), pioneered by Ho et al. \cite{ho2020denoising}, represent a significant advancement in generative modeling, demonstrating superior sample quality compared to GANs for both unconditional image generation and image-to-image tasks \cite{dhariwal2021diffusion, song2020denoising, nichol2021improved}. In computational pathology, DDPMs have enabled the generation of high-fidelity synthetic images across various applications, including H\&E images, rare cancer datasets, and gigapixel pathology images \cite{moghadam2023morphology, oh2023diffmix, shrivastava2023nasdm, kataria2024staindiffuser, yu2023diffusion, yellapragada2024pathldm, harb2024diffusion, aversa2024diffinfinite,winter2025utilizing}.
The introduction of conditional DDPMs enabled targeted data generation by constraining the inherently stochastic diffusion process through various guidance mechanisms \cite{nichol2021glide}.
Building upon conditional DDPMs, semantically conditioned diffusion models (SDMs) \cite{wang2022semantic} have advanced guided medical image generation, particularly in brain MRI \cite{hung2023med} and H\&E image synthesis \cite{xu2025topocellgen}. These models leverage high-level contextual cues through semantic label maps for targeted synthesis of anatomical structures.
The SDM architecture employs a U-Net-based conditional denoising network that processes noisy images through encoder blocks while injecting semantic label maps into decoder blocks via spatially-adaptive normalization (SPADE) \cite{park2019semantic}. The encoder incorporates timestep-dependent scaling through specialized residual blocks, while the decoder uses SPADE to enable spatially-adaptive feature modulation.
Parallel to semantic conditioning, text-conditioned diffusion models guide synthesis using natural language prompts \cite{avrahami2022blended}, finding applications in biomedical image generation for H\&E slides \cite{rao2024improving} and spatially consistent nuclei generation \cite{oh2024controllable}.
However, applying conventional SDMs to computational pathology presents unique challenges. Addressing pathology-specific challenges, namely that pathological images contain numerous, small, often touching cell instances, necessitates more fine-grained spatial conditioning for enhanced instance separation, which conventional semantic conditioning approaches do not fully address. Our work extends these conditioning mechanisms by integrating multimodal guidance to meet the specific requirements of pathological image synthesis.
\section{Methods}
\subsection{MSDM: Extending the semantic diffusion model}
We re-implemented the SDM \cite{wang2022semantic} based on the guided diffusion model \cite{dhariwal2021diffusion} to meet the requirements of computational pathology by first adapting the semantic maps that guide the diffusion process to horizontal and vertical (HV) maps. 
These HV maps are calculated as the horizontal/vertical distances to the boundaries of each instance (e.g., cell/nuclei boundaries), enhancing instance separation.
We extend the conditioning with RGB color characteristics, creating a 16-channel guidance maps.
For both cells and nuclei, we include HV maps (2 channels), mean RGB values (3 channels), and RGB standard deviations (3 channels), totaling 8 channels for cells and eight channels for nuclei. These maps display one constant value for the mean and standard deviation of all cells and for the mean and standard deviation of all nuclei in a given image. 
This 16-channel mask guides the diffusion process via SPADE normalization in the U-Net decoder (see figure \ref{figure1}) \cite{park2019semantic}.
To further enhance guidance, we add text conditioning using BERT-tiny \cite{devlin2019bert} to encode assay and indication information. Multi-head cross-attention in the diffusion model's U-Net bottleneck fuses encoder features with text tokens. The bottleneck was chosen for its semantically rich image representation, maximal global context, and computational efficiency due to the most feature compression.

\subsection{Multimodal feature fusion}
Multimodal feature fusion employs multi-head cross-attention, integrating image and text features for semantically and textually guided image generation.
The latent representations of the current noisy image (image features) and the encoded text embeddings are fused using cross-attention. 
Given the image feature matrix ($X \in \mathbb{R}^{H \times W \times C}$) and the text feature matrix ($T \in \mathbb{R}^{L \times D}$), the model projects these to query, key, and value spaces as follows: $Q = X W_Q$, $K = T W_K$, $V = T W_V$
where ($W_Q$, $W_K$, and $W_V$) are learnable projection matrices. For each attention head, the cross-attention is computed as:
\begin{equation}
  \text{Attention}(Q, K, V) = \text{softmax}\left( \frac{Q K^{T}}{\sqrt{d}} \right) V
\end{equation}
Here, queries come from image features and keys/values from text, allowing the model to align and integrate textual context at each spatial location. We run eight attention heads $i \in \{0, \dots, 7\}$ in parallel, each capturing different image–text correlation patterns. 
\begin{align}
  \text{MultiHeadAttention}(Q, K, V) =
  & \\ \big( \text{Concat}_{i=1}^{h} \left( 
      \text{Attention}(Q_i, K_i, V_i)
     \right)
  \nonumber
  &\quad \times W^{O} \big) + b^{o}
\end{align}
The attended feature vector is added to the image feature vector. This fused representation is propagated through the diffusion denoising steps, allowing the model to iteratively refine image predictions in a way that accurately reflects the provided textual guidance.

Latent space features were investigated by embedding real and synthetic images using the H-Optimus-0 foundation model \cite{saillard2024hoptimus}, visualizing embeddings with UMAP \cite{mcinnes2018umap}, and calculating Wasserstein distances \cite{vaserstein1969markov} between distributions.
\section{Experiments}
We compare the baseline Semantic Diffusion Model (SDM), only modified with 4-channel HV spatial conditioning for cell and nuclei specificity with our MSDM.

\subsection{Model training}
Both diffusion models for data augmentation were trained using the ADAM optimizer with mean squared error loss and a cosine noise scheduler for 20,000 epochs. A batch size of 5 with 5 gradient accumulation steps yielded an effective batch size of 25. During inference, we used DDIM scheduler \cite{song2020denoising} with 40 timestep respacing steps, yielding 25-fold faster inference versus DDPM scheduler.
The validation set remained constant for all trainings.
\subsection{Datasets}
Our training dataset comprises 566 immunohistochemistry (IHC)-stained slides (e.g., HER2 membrane marker assay, 20x magnification). From these, 612 Field-of-Views (FOVs) of size 512x512 pixels were selected. 
A group of board-certified pathologists annotated all cell and nuclei outlines within the tumor epithelium of these FOVs, yielding 307,666 cell and 232,031 nuclei outlines for the training set.
The manually annotated validation dataset was established using 267 independent IHC-stained slides, also featuring the same assay. A total of 309 FOVs of size 512x512 pixels were selected from these slides, and all cell and nuclei outlines within them were annotated by a single pathologist. This resulted in 153,873 annotated cell outlines and 110,834 annotated nuclei outlines.
We have two test sets, the first containing mainly columnar cells and the second containing a mix of indications and cells. 
Our first test set contains 36 FOVs (512×512 pixel) from 29 IHC slides; the second has 303 FOVs from 147 slides representing diverse indications and assays, with no particular focus on columnar cells. Within the tumor epithelium, cell centers were independently annotated by three pathologists for all FOVs, and cell/nuclei outlines for selected cells. First test set annotation counts (P1/P2/P3): cell centers — 12869/12741/14186; cell outlines — 933/698/698; nuclei outlines — 898/699/698. Second test set: cell centers — 196825/12741/14186; cell outlines — 149074/5672/6177; nuclei outlines — 155560/5668/6219. Only tumor epithelium regions were annotated and analyzed.
\subsection{Dataset enhancement with the diffusion model}
\label{sec:enhancement}
To focus our augmentation efforts on columnar cells, we identified FOVs in our training dataset containing cells with high eccentricity.
Cell eccentricity was calculated by fitting an ellipse to the cell shape and determining its eccentricity $e$ from the minor/major axes ratio:
\begin{equation}
 e = \sqrt{1 - \left(\frac{L_{\text{minor}}}{L_{\text{major}}}\right)^2}
\label{eq:eccentricity} 
\end{equation}
\begin{figure}[htb]
  \begin{minipage}[b]{1.0\linewidth}
    \centering
    \centerline{\includegraphics[width=8.5cm]{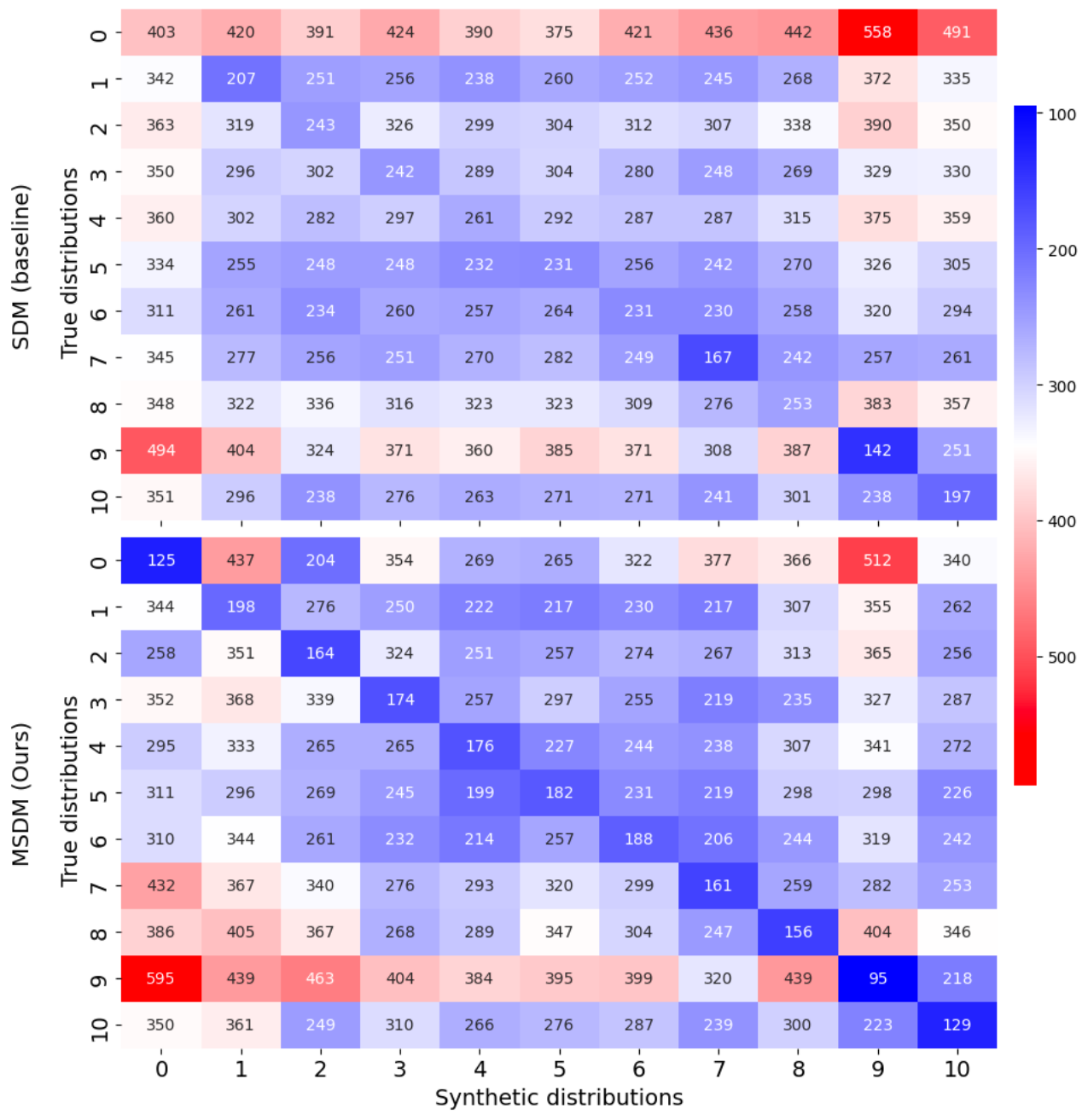}}
    \vspace*{-2mm}
    \caption{
      Wasserstein distances in latent space embeddings between real (y-axis) and synthetic (x-axis) images corresponding to different combination of assay and indication pairs (e.g. 1:=HER2/Breast,...). Matching pairs lie on the diagnonal of the so-formed matrices.
    }
    \vspace*{-2mm}
  \label{figure3}
\end{minipage}
\end{figure}
FOVs with an average cell eccentricity $e>0.85$ were chosen, yielding 120 FOVs (of the 612 FOVs in the training dataset). For image generation, one cell mask was randomly chosen from the 120 chosen FOVs, together with its corresponding nuclei mask. Color and metadata conditions were randomly selected from the 120 chosen FOVs. Choosing newly matched condition pairs for each image generation, we generated eight new versions for each FOV, resulting in 960 synthetic FOVs (see figure \ref{figure2}). Re-using existing masks, color, and metadata ensured synthetic images were realistic examples, reflecting desired morphological properties.
\subsection{Evaluation}
To assess the impact of our synthetic data in downstream segmentation tasks, Cellpose \cite{stringer2021cellpose} models were trained for cell and nuclei segmentation under three distinct scenarios, all utilizing the same validation set for consistent evaluation. These scenarios include: (1) training with all 612 annotated real training data only (serving as our baseline); (2) training with all 612 annotated real training data plus 960 synthetic images created with the SDM model; and (3) training with all 612 annotated real training data plus 960 synthetic images created with our MSDM model. The synthetic data generated by SDM and MSDM was combined with existing image-mask pairs in the training dataset (see figure 2).
The segmentation models were benchmarked against three pathologists' annotations on the test set using F1 score for cell center detection, average symmetric surface distance (ASSD) for membrane segmentation, and Dice score for cytoplasm and nuclei segmentation (after Hungarian matching). 
We prioritized cell center detection (F1 score) for evaluating performance because other metrics (membrane, cytoplasm, nuclei segmentation) are dependent on successful cell center identification via Hungarian matching.
For final results (see Table \ref{tab:Results}), each model training was repeated five times, evaluated against all three pathologists, and metrics were averaged.
To statistically assess the significance of observed differences in performance between models, we employed a Wilcoxon signed-rank test, considering a difference significant when the p-value was less than $p<0.05$.
Beyond assessing the utility of synthetic data for downstream tasks, we evaluated the faithfulness of the synthetic data. 
Therefore, we calculated the Wasserstein distances in latent space embeddings between indication-assay pairs of real and synthetic data contained in the validation set (see Figure \ref{figure3}). 
This approach provides insight into how closely the embedding space of the synthetic data aligns with that of the real data, thereby indicating how well the generative models capture the underlying data distribution.

\section{Results}
MSDM demonstrates effectiveness through two key evaluations.
First, we verify conditioning efficacy by generating images guided by textual inputs (indications and assays). Using metadata, we identified ten distinct indication-assay pairs as clusters for evaluating conditioning. We used H-Optimus-0, a state-of-the-art computational pathology foundation model, to embed synthetic images generated from these pairs and corresponding real validation images. In the embedding space, we computed Wasserstein distances between real and synthetic data for each of the ten indication-assay pairs.
MSDM achieved significantly lower average Wasserstein distances ($158.75 \pm 29.50; mean \pm sdev$) for matching distributions and for for non-matching distributions ($303.46 \pm 69.73$) (see Fig. \ref{figure3}) than the SDM model (matching: $234.18 \pm 63.92$; non-matching: $312.87 \pm 61.95$).
Second, we investigated performance improvements in columnar cell and nuclei segmentation using our synthetic data. MSDM yielded best results (see Table \ref{tab:Results}), significantly outperforming models trained solely on real data and surpassing the SDM model. On our second test set containing mixed cells, our approach increases cytoplasm and nuclei segmentation performance while retaining cell center detection and membrane segmentation performance.
\begin{table}[h]
  \centering
  \begin{tabular}{|p{0.93cm}|p{1.34cm}|p{1.34cm}|p{1.34cm}|p{1.34cm}|}
    \hline
    \multicolumn{5}{|c|}{\textbf{First test set (Columnar cells)}} \\
    \hline
    Results & Cell center detection & Membrane segmentation & Cytoplasm segmentation & Nuclei segmentation \\
    \hline
    Metric & F1 [$\%$] \newline $\uparrow$ better & ASSD \newline $\downarrow$ better & \multicolumn{2}{|>{\centering}p{2.6cm}|}{Dice score [$\%$] \newline $\uparrow$ better}\\
    \hline
    IPA & 86.62$\pm$0.82& 0.62$\pm$0.02 & 72.18$\pm$1.42 & 90.32$\pm$0.63 \\
    \hline
    Baseline & 81.99$\pm$0.77 & 0.99$\pm$0.04 & 63.68$\pm$1.37 & 81.52$\pm$1.29 \\
    \hline
    SDM & 81.64$\pm$0.74 $\star$ & 0.97$\pm$0.04 $\star$ & \textbf{64.36$\pm$1.39} $\star$ & 82.14$\pm$1.35 $\star$ \\
    \hline
    MSDM & \textbf{82.30$\pm$0.87} $\star \times$ & \textbf{0.96$\pm$0.04} $\star$ & 64.19$\pm$1.47 $\star$ & \textbf{82.23$\pm$1.37} $\star$ \\
    \hline
    \multicolumn{5}{|c|}{\textbf{Second test set (Mixed cells)}} \\
    \hline
    IPA & 86.84$\pm$2.66 & 0.70$\pm$0.13 & 71.10$\pm$5.14 & 89.49$\pm$2.50 \\
    \hline
    Baseline & 82.53$\pm$4.11 & 1.00$\pm$0.26 & 62.93$\pm$7.50 & 84.41$\pm$4.64 \\
    \hline
    MSDM &  \textbf{82.90$\pm$3.62} &  \textbf{0.99$\pm$0.22} & \textbf{64.31$\pm$5.21} $\star$ & \textbf{85.42$\pm$3.95} $\star$ \\
    \hline
  \end{tabular}
  \caption{Reported values are means and standard deviation. IPA: agreement among three pathologists. Significant differences ($p<0.05$): $\star$ (baseline vs. SDM/MSDM), $\times$ (SDM vs. MSDM).}
\label{tab:Results}
\end{table}

\section{Discussion}
In this study, we modified a semantic diffusion model, creating a novel multimodal-conditioned generative framework for computational pathology. 
Our architectural enhancements, incorporating HV maps for morphology, RGB color, and BERT-encoded metadata, enable the generation of targeted synthetic data to specifically enhance the segmentation performance of morphologies with low prevalence, exemplified on columnar cells. 
The observed Wasserstein distances between embedded real and synthetic data for matching biological conditions demonstrate our model's capacity to faithfully capture crucial morphological and contextual nuances, as compared to the SDM.
In a downstream application, the synthetically generated images significantly boost the performance of cell and nuclei segmentation models on IHC-stained images, yielding superior performance as compared to models trained on real data alone or data generated with the SDM model.
In comparison to manually annotating FOVs by a pathologists, which we estimate to take around 4 hours per FOV (accumulating to 3840 hours for 960 FOVs) this presents a significant time and cost saving, although one might argue that FOVs placed and annotated by pathologists might be of larger benefit for model training.
We acknowledge that, by reusing existing masks, our current approach for synthetic data generation requires an initial set of annotations on cells with challenging morphologies. 
This implies that while our method effectively augments existing datasets, it does not fully circumvent the need for initial annotations.
In future work, we plan to explore broader applications, including other challenging morphologies, diverse assays, and potential computational efficiencies, to fully leverage the power of multimodal diffusion models for data augmentation in computational pathology. 
This study thus underscores the potential of generative models in overcoming data scarcity and advancing model robustness for critical pathology tasks. 
\section{Compliance with ethical standards}
\label{sec:ethics}
The study was conducted in adherence to the International Council for Harmonization Good Clinical Practice guidelines, the Declaration of Helsinki, and local regulations on the conduct of clinical research. The AstraZeneca Biobank in the UK is licensed by the Human Tissue Authority (License No. 12109) and has National Research Ethics Service Committee (NREC) approval as a Research Tissue Bank (RTB) (REC No 17/NW/0207) which covers the use of the samples for this project. The studies were conducted in accordance with the local legislation and institutional requirements. The participants provided their written informed consent to participate in this study.
\section{Acknowledgments}
\label{sec:acknowledgments}
All authors are employees of AstraZeneca and do not have other relevant financial or non-financial interests to disclose. We thank Helen Angell, Mari Heininen-Brown, and Andrea Storti on behalf of the Tumour \& Immune Cell Atlas team (TICA, AstraZeneca), whose end-to-end stewardship of a large, rigorously curated histological slide database (sourcing, staining, digitalisation, curation and QA) enabled the training of our diffusion models. 

% Below is an example of how to insert images. Delete the ``\vspace'' line,
% uncomment the preceding line ``\centerline...'' and replace ``imageX.ps''
% with a suitable PostScript file name.
% -------------------------------------------------------------------------


\bibliographystyle{IEEEbib}
\bibliography{bibliography}
\end{document}